\documentclass[letterpaper, 10 pt, conference]{IEEEtran}
\usepackage[protrusion=true,expansion=false]{microtype}

\IEEEoverridecommandlockouts

\usepackage{gensymb}
\usepackage{amsmath,amssymb,amsfonts}
\usepackage{algorithmic}
\usepackage{graphicx}
\usepackage{textcomp}
\usepackage{xcolor}
\usepackage{booktabs}
\usepackage{hyperref}
\hypersetup{colorlinks=true, linkcolor= blue, allcolors=blue, citecolor = blue, filecolor=black, urlcolor=blue}
\usepackage{float}
\usepackage{comment}
\usepackage{mathtools}
\usepackage{subcaption}
\usepackage{caption}
\def\BibTeX{{\rm B\kern-.05em{\sc i\kern-.025em b}\kern-.08em
  T\kern-.1667em\lower.7ex\hbox{E}\kern-.125emX}}

\usepackage{siunitx} 
\usepackage{tikz}
\usetikzlibrary{arrows.meta}
\usepackage{mylatexdefs}

\newcommand{\bbsym}[1]{\ensuremath{\boldsymbol{#1}}}

\begin{document}

\title{\LARGE \bf Nonplanar Model Predictive Control for Autonomous Vehicles\\ with Recursive Sparse Gaussian Process Dynamics}

\author{Ahmad Amine, 
Kabir Puri, Viet-Anh Le,
and Rahul Mangharam
\thanks{The authors are with Department of Electrical \& Systems Engineering, University of Pennsylvania, Philadelphia, PA 19104, USA (emails: {\tt\small \{aminea,kapuri,vietanh,rahulm\}@seas.upenn.edu}).}
}

\author{
	\parbox{\textwidth}{%
		\centering
		Ahmad Amine$^{1}$, Kabir Puri$^{1}$, Viet-Anh Le$^{1}$, Rahul Mangharam$^{1}$%
	}%
	\thanks{$^{1}$Department of Electrical \& Systems Engineering, University of Pennsylvania, Philadelphia, PA 19104, USA
		{\tt\small aminea@seas.upenn.edu, kapuri@seas.upenn.edu, vietanh@seas.upenn.edu, rahulm@seas.upenn.edu}}
    \thanks{This work was partially supported by US DoT Safety21 National University Transportation Center and NSF under Grant CISE-$2431569$.}
    \thanks{{Paper Website: \url{https://sites.google.com/seas.upenn.edu/npmpc/home}}}
}
\maketitle

\begin{abstract}
This paper proposes a nonplanar model predictive control (MPC) framework for autonomous vehicles operating on nonplanar terrain. To approximate complex vehicle dynamics in such environments, we develop a geometry-aware modeling approach that learns a residual Gaussian Process (GP). By utilizing a recursive sparse GP, the framework enables real-time adaptation to varying terrain geometry. The effectiveness of the learned model is demonstrated in a reference-tracking task using a Model Predictive Path Integral (MPPI) controller. Validation within a custom Isaac Sim environment confirms the framework's capability to maintain high tracking accuracy on challenging 3D surfaces.
\end{abstract}

\section{Introduction}

Although extensive research efforts have addressed autonomous driving tasks, prior research has predominantly focused on flat terrain conditions \cite{betz2022autonomous}.
However, in many real-world applications, autonomous vehicles must be able to operate in unstructured and off-road environments.
In particular, off-road driving requires safe navigation over nonplanar or uneven terrain, including slopes, hills, banks, and bumps.
Navigating autonomous vehicles in such terrain environments requires a precise understanding of 3D vehicle dynamics, which can then be incorporated into model predictive control.

Recent work has addressed dynamic modeling and navigation by accounting for terrain topology of nonplanar surfaces.
Some studies on nonplanar modeling considered the road as a ribbon, and developed the dynamics of the vehicle based on the assumption that the vehicle moves on Darboux tangent frame at the nearby spine point.
The idea was first introduced by Limebeer \& Perantoni \cite{limebeer2015optimalpart2}.
Fork \& Borrelli in \cite{fork2021models,fork2024models} used Tait-Bryan (TB) angles along the trajectory for surface parameterization, and derived a kinematic model for the vehicle on the nonplanar parametric surface.
Yu \etal \cite{yu2021nonlinear} presented an extended bicycle model that takes into account the topology of the terrain through a rotation transformation using the TB angles.
Piccinini \etal \cite{piccinini2025kineto} developed an artificial race driver that integrates a kineto-dynamical vehicle model to learn the vehicle dynamics, plan, and execute minimum-time maneuvers on a 3D track.
Other studies investigated planning directly on point cloud maps.
For example, Kr{\"u}si \etal \cite{krusi2017driving} introduced a motion planning framework on generic 2D manifolds embedded in the 3D space without requiring explicit topology extraction.
Han \etal \cite{han2023model} proposed a physics-based framework to derive traversability constraints, and a planning framework that generates sampled rollouts on a 2D kinematic plane, then projects the rollouts on a 3D elevation map.
Datar \etal \cite{datar2024learning} proposed a learning framework to model the forward vehicle-terrain dynamics, in which the terrain's elevation map is incorporated as part of the neural network input.
Lee \etal \cite{lee2023learning} proposed a terrain-aware kinodynamic model, which combined an elevation map encoder, a dynamics predictive neural network, and an explicit kinematic layer.

Due to the difficulty in analytically modeling complex vehicle dynamics, and the effects of the terrain geometry and interaction, learning-based nonplanar vehicle modeling has been considered in recent work \cite{datar2024learning,lee2023learning}, which have used neural networks.
However, neural networks require large training data, and the training process must be performed offline.
As a result, the model lacks the ability to adapt online to capture the uncertainty of the varying terrain.
Therefore, in this work, we propose a new approach to approximate the dynamics of an autonomous vehicle on nonplanar terrain in real-time.
In particular, our approach relies on a sparse Gaussian process (GP) model with online recursive updates.
In the proposed geometry-aware model, we combine a nominal single-track dynamic model with a data-driven residual model using a sparse GP that takes into account the geometry of the terrain. 
The sparse GP model is updated recursively given real-time collected data.
The proposed model is then used in a model predictive control (MPC) formulation, and a model predictive path integral (MPPI) framework~\cite{williams2017information} is used to solve the complex MPC problem to generate the optimal control inputs in a receding horizon manner.
We validate the proposed framework in a simulation environment where we generate different nonplanar tracks with slopes, hills, banks, and bumps.

The rest of this paper is organized as follows.
In Section~\ref{sec:model}, we present the vehicle model on a nonplanar surface. 
In Section~\ref{sec:npmpc}, we formulate the MPC problem and how to solve the problem with MPPI.
In Section~\ref{sec:sim}, we show the simulation results in the \texttt{Isaac Sim} environment, and we provide some concluding remarks in Section~\ref{sec:cls}.

\begin{figure*}[t]
\centering
\begin{subfigure}{0.33\textwidth}
\centering
\includegraphics[width=\linewidth,trim=0 0 0 40,clip]{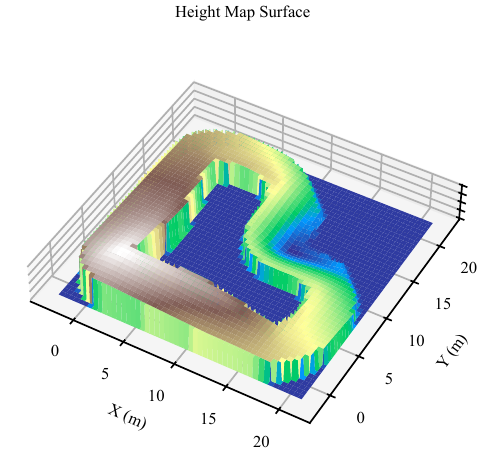}
\vspace{-5pt}
\caption{3D height map}
\label{fig:height}
\end{subfigure}
\hfill
\hspace{-10mm}
\begin{subfigure}{0.33\textwidth}
\centering
\includegraphics[width=\linewidth,trim=0 0 0 0,clip]{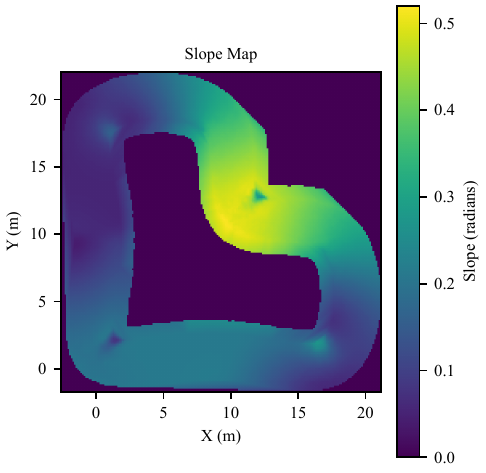}
\vspace{-5pt}
\caption{Slope map}
\label{fig:slope}
\end{subfigure}
\begin{subfigure}{0.33\textwidth}
\centering
\includegraphics[width=\linewidth,trim=0 0 0 0,clip]{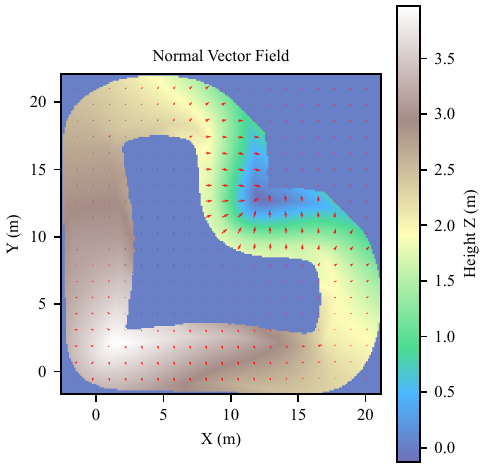}
\vspace{-5pt}
\caption{Normal vector field}
\label{fig:vector}
\end{subfigure}
\caption{The elevation map of the terrain for a specific track.}
\label{fig:slopemap}
\vspace{-6mm}
\end{figure*}

\section{Learning-Based Nonplanar Vehicle Model}
\label{sec:model}

In this section, we first present the problem of modeling vehicle dynamics on a nonplanar terrain and then develop a learning-based model using sparse GPs with recursive updates \cite{liu2025learning}.
We consider the motion of a ground vehicle traveling over a nonplanar surface. 
We denote the vectors of states and control inputs of the vehicle as $\bbsym{x}$ and $\bbsym{u}$, respectively.
We define the state vector as
$\bbsym{x} = [p_x, p_y, \psi, \delta, v, \beta, r]^\top \in \mathbb{R}^7$,
where $[p_x,p_y]^\top$ are the $x$ and $y$ positions, $\psi$ is the yaw angle, $\delta$ is steering angle, $v$ is the velocity in car frame, $\beta$ is side-slip angle, and $r=\dot{\psi}$ is yaw rate. The vector of control inputs is $\bbsym{u} = [a, v_{\delta}]^\top \in \mathbb{R}^2$ that includes
the longitudinal acceleration $a$ and steering-rate $v_{\delta}$.


\subsection{Terrain Modeling}
\label{sec:terrain_modeling}

We model the terrain as a smooth height field $h:\mathbb{R}^2 \to \mathbb{R}$ over a global frame in which the $z$-axis is vertical and the $(x,y)$ plane is the nominal ground plane. 
At each planar location $(p_x,p_y)$, we associate a unit surface normal $\bbsym{n}(p_x,p_y) = [n_x, n_y, n_z]^\top$, a scalar slope angle $\theta(p_x,p_y)$, and a surface orientation angle $\varphi(p_x,p_y)$. 
We denote $\TTT (p_x, p_y)$ as the tangent plane at the vehicle's location $(p_x, p_y)$. 
The normal $\bbsym{n}(p_x,p_y)$ is defined as the unit vector orthogonal to the local tangent plane and oriented so that $\bbsym{n}(p_x,p_y) \cdot \bbsym{e}_z \ge 0$, fixing ``upwards'' along $+z$-axis. 
The slope angle is the inclination of the surface with respect to the vertical,
\[
\theta(p_x,p_y) = \arccos\bigl(|n_z(p_x,p_y)|\bigr)
\]
The surface orientation is defined as the azimuth of the projection of $\bbsym{n}(p_x,p_y)$ onto the $(x,y)$ plane,
\[
\varphi(p_x,p_y) = \operatorname{atan2}\bigl(n_y(p_x,p_y),\,n_x(p_x,p_y)\bigr),
\]
and the associated unit vector $(\cos\varphi(p_x,p_y),\sin\varphi(p_x,p_y))$ in the horizontal plane is interpreted as pointing in the direction of steepest ascent along the surface.

We start from a triangulated surface mesh $\MMM \subset\mathbb{R}^3$ obtained by an offline 3D mapping pipeline (e.g., from LiDAR or depth-based reconstruction) or constructed for simulation. 
Points $\{\bbsym{p}_k\}_{k=1}^N$ are sampled on $\MMM$ using Poisson disk sampling, and per-point normals $\{\bbsym{n}_k\}_{k=1}^N$ are estimated by local tangent plane fitting, with consistent orientation enforced and oriented so that $\bbsym{n}_k\cdot \bbsym{e}_z \ge 0$. 
These operations are implemented using Open3D~\cite{Zhou2018} in Python. Writing $\bbsym{p}_k=(x_k,y_k,z_k)$, we construct an axis-aligned bounding box in the $(x,y)$ plane and a regular Cartesian grid $\{(x_i,y_j)\}_{i,j}$. 
The height field $h$, the components of the normal field $\bbsym{n}$, and the derived slope and orientation angles $(\theta,\varphi)$ are then obtained by interpolating $\{(x_k,y_k,z_k,\bbsym{n}_k)\}$ onto this grid using linear interpolation. 
The same construction could be applied in a robot-centric frame using the current local 3D scan to build a moving grid of $h$, $\bbsym{n}$, $\theta$, and $\varphi$ for mapless control.
In this paper, we focus on the offline, globally mapped setting. 
We illustrate the elevation map, including a height map, a slope map, and a normal vector field, of the terrain for a specific track in Fig.~\ref{fig:slopemap}.

\subsection{Learning-Based Nonplanar Dynamics}
\label{sec:model-b}
Next, we present our approach to approximate the vehicle model on nonplanar terrain based on combining a single-track bicycle model and a GP model for learning the residual dynamics.
The single-track bicycle model describes the vehicle dynamics on a planar surface, thus we utilize a GP to learn the residual dynamics due to the nonplanar effects. 

The single-track bicycle model is regularly used to model the vehicle dynamics on a planar surface.
The full details of the single-track dynamic model can be found in \cite{althoff2017commonroad}, while we represent it in the following compact form
\begin{equation}
\label{eq:st_model}
\frac{d \bbsym{x}}{dt} 
=
\bbsym{f}_\mathrm{ST} (\bbsym{x}, \bbsym{u}\big).
\end{equation}
The continuous model \eqref{eq:st_model} can be discretized using an ordinary differential equation (ODE) solver, leading to the following discrete-time model:
\begin{equation}
\label{eq:dis_st_model}
\bbsym{x}_{k+1} 
=
\mathrm{ODESolve}\big(\bbsym{f}_\mathrm{ST}, \bbsym{x}_{k}, \bbsym{u}_{k}\big),
\end{equation}
where $k$ is the time step.

Since the single-track dynamic model \eqref{eq:dis_st_model} does not accurately capture the vehicle dynamics on a nonplanar surface, we propose to compensate by a residual learning-based model.  

Let $\alpha (p_x, p_y)$ and $\gamma (p_x, p_y)$ be the roll and pitch angles to obtain the tangent frame $\TTT (p_x, p_y)$ from the global frame. 

Given a normal vector $\bbsym{n} (p_x, p_y) = [n_x, n_y, n_z]^\top$, the roll and pitch angles
can be computed as
\begin{subequations}
\begin{align}
\alpha (p_x, p_y) &= \arctan2(n_y,\, n_z), \\[2pt]
\gamma (p_x, p_y) &= \arctan2\!\left(-n_x,\; \sqrt{n_y^2 + n_z^2}\right).
\end{align} 
\end{subequations}
We learn residual dynamics for $[v, \beta, r]$ using a sparse GP conditioned on terrain geometry (roll and pitch). The GP is updated online at each time step via recursive updates.
Let $\GGG\PPP (\bbsym{\xi}): \RR^9 \to \RR^3$ be the multi-output sparse GP model for the residual dynamics, where the vector of inputs for the sparse GPs is defined as follows,
\begin{equation}
\bbsym{\xi} = [\psi, \delta, v, \beta, r, a, v_{\delta}, \alpha, \gamma]^\top.
\end{equation}
Note that in this paper, we approximate a multi-output GP by multiple single-output GPs. This formulation allows the learned residual to be predicted at future states obtained by propagating the nominal dynamics forward. At each future state, we query the GP with the map-derived terrain geometry and the predicted state to produce a terrain-topology conditioned compensation to the dynamics. As a result, the nonplanar vehicle model is approximated by the single-track model and a residual sparse GP as follows,
\begin{equation}
\bbsym{x}_{k+1} = \mathrm{ODESolve}\big(\bbsym{f}_\mathrm{ST}, \bbsym{x}_{k}, \bbsym{u}_{k}\big)
+ \bbsym{H}\, \GGG\PPP (\bbsym{\xi}_k ),
\end{equation}
where $\bbsym{H} = [\bbsym{0}_{3\times 4}, \II_{3}]^\top \in \RR^{7 \times 3}$.
The details on recursive updates for sparse GPs are given in the next section.

\subsection{Sparse Gaussian Processes with Recursive Updates}

We utilize sparse Gaussian processes (GPs) with online recursive updates to continuously learn the unknown time-varying uncertainties in the vehicle dynamics and the terrain.
A sparse GP approximates a full GP by introducing a finite set of inducing points, which reduces the computational cost of standard GPs.
Moreover, since the weight-space representation of a sparse GP can be written as a linear combination of kernel functions centered at the inducing points, the model parameters can be updated efficiently using a recursive least-squares algorithm~\cite{liu2025learning}. 

Let $\DDD_N = \{\bbsym{Z}, \bbsym{Y}\}$ denote the training dataset, where $\bbsym{Z}$ and $\bbsym{Y}$ are the training inputs and outputs, respectively, and let $\bbsym{Z}_\mathrm{u}$ be a set of pseudo-inputs (inducing points).
In a variational sparse GP, the true posterior distribution $P(\bbsym{\Delta} \mid \DDD_N)$ is approximated by
\(
q(\bbsym{\Delta},\bbsym{\Delta}_\mathrm{u}) 
= P(\bbsym{\Delta} \mid \bbsym{\Delta}_\mathrm{u})\, q(\bbsym{\Delta}_\mathrm{u}),
\)
where $q\left(\bbsym{\Delta}_\mathrm{u}\right) = \NNN(\bbsym{\Delta}_\mathrm{u} \mid \bbsym{m}_\mathrm{u}, \bbsym{S}_\mathrm{u})$ is a tractable variational distribution obtained by maximizing the evidence lower bound (ELBO) of $\log P(\bbsym{Y})$.
The mean and covariance of $q\left(\bbsym{\Delta}_\mathrm{u}\right)$ are given by
\begin{subequations}
\begin{align}
\bbsym{m}_\mathrm{u} &= \sigma_{\epsilon}^{-2}\,\bbsym{S}_\mathrm{u} \bbsym{K}_{M}^{-1} \bbsym{K}_{MN} \bbsym{Y}, \\[1mm]
\bbsym{S}_\mathrm{u} &= \bbsym{K}_{M}\left(\bbsym{K}_{M} + \sigma_{\epsilon}^{-2} \bbsym{K}_{MN} \bbsym{K}_{NM}\right)^{-1} \bbsym{K}_{M},
\end{align}
\end{subequations}
where $[\bbsym{K}_{MN}]_{i,j} = \kappa(\bbsym{z}_{\mathrm{u},i}, \bbsym{z}_{j})$ is the covariance matrix between the inducing inputs $\bbsym{Z}_\mathrm{u}$ and the training inputs $\bbsym{Z}$, $\bbsym{K}_{NM} = \bbsym{K}_{MN}^{\top}$, and $\bbsym{K}_{M}$ is the covariance matrix of the inducing inputs.

As a result, at a test point $\bbsym{z}^{*}$, the predictive mean and variance are
\begin{subequations}
\label{eq:sgp-pred}
\begin{align}
\label{eq:sgp-pred-a}
\mu_{\Delta}(\bbsym{z}^{*}) &= \bbsym{K}_{*M} \bbsym{K}_{M}^{-1} \bbsym{m}_\mathrm{u}, \\
\label{eq:sgp-pred-b}
\sigma^2_{\Delta}(\bbsym{z}^{*}) &= k_{**} - \bbsym{K}_{*M}\!\left(\bbsym{K}_{M}^{-1} - \bbsym{K}_{M}^{-1} \bbsym{S}_\mathrm{u} \bbsym{K}_{M}^{-1}\right) \bbsym{K}_{M*},
\end{align}
\end{subequations}
where $[\bbsym{K}_{*M}]_{j} = \kappa(\bbsym{z}^{*}, \bbsym{z}_{\mathrm{u},j})$, $\bbsym{K}_{M*} = \bbsym{K}_{*M}^{\top}$, and $k_{**} = \kappa(\bbsym{z}^{*}, \bbsym{z}^{*})$.

The sparse GP model can be updated recursively online as new data are collected.
Given a new online output at time step $k$, the posterior mean and variance for $q_k (\bbsym{\Delta}_{\mathrm{u}})$ can be computed by 
\begin{subequations}
\begin{align}
&\bbsym{m}_\mathrm{u}^{k} = \bbsym{S}_\mathrm{u}^{k}(\bbsym{S}_\mathrm{u}^{k} \bbsym{m}_\mathrm{u}^{k} 
+ \sigma_{\epsilon}^{-2}\bbsym{\Phi}_k^{\top}{y}(k)), \\
&\bbsym{S}_\mathrm{u}^{k} = (\bbsym{S}_\mathrm{u}^{k-1} 
+ \sigma_{\epsilon}^{-2} \bbsym{\Phi}_k^{\top} \bbsym{\Phi}_k)^{-1},
\end{align}
\end{subequations}
where kernel $\bbsym{\Phi}_k=\bbsym{K}_{zM} \bbsym{K}_{M}^{-1}|_{{\bbsym{z}}=\bbsym{z}(k)}$ and $[\bbsym{\Phi}({\bbsym{z}}(k))]_{j} = \phi{\bbsym{z}}(k),{\bbsym{z}}_\mathrm{u}(j))$. 
Then \eqref{eq:sgp-pred-a} can be seen as a linear combination of $M$ kernel functions.
Thus, recursive least squares can be efficiently used to update~\eqref{eq:sgp-pred-a} and~\eqref{eq:sgp-pred-b} online. 
Thus, given a new data point $\{ \bbsym{z} (k), {y}(k) \}$, the posterior mean and variance can be updated by:
\begin{equation}
\begin{split}
\bbsym{m}_\mathrm{u}^{k} &=\bbsym{m}_\mathrm{u}^{k-1}+\bbsym{L}_k{r}_k, \;
{r}_k = {y}(k) - \bbsym{\Phi}_k \bbsym{m}_\mathrm{u}^{k-1} \\
\bbsym{L}_{k} &= \bbsym{S}_\mathrm{u}^{k-1} \bbsym{\Phi}_k^{\top} \bbsym{G}_{k}^{-1}, \; \bbsym{G}_{k} = \lambda + \bbsym{\Phi}_k \bbsym{S}_\mathrm{u}^{k-1} \bbsym{\Phi}_k^{\top} \\
\bbsym{S}_\mathrm{u}^{k} &= \lambda^{-1}(\bbsym{S}_\mathrm{u}^{k-1} -\bbsym{L}_{k}\bbsym{G}_{k}\bbsym{L}_{k}^{\top}),
\end{split}
\end{equation}
where $0<\lambda\leq 1$ is the forgetting factor.
The recursion starts from $\bbsym{m}_\mathrm{u}^{0}$ and $\bbsym{S}_\mathrm{u}^{0}$, which can be seen as the prior of the GP.
Note that the sparse GP must be trained on some offline training dataset to obtain initial hyperparameters of the GP kernel that are fixed during online recursive updates.

\section{Nonplanar Model Predictive Control}
\label{sec:npmpc}

In this section, we formulate a nonplanar model predictive control (MPC) tracking problem using the learning-based nonplanar vehicle model presented in the previous section. 

\subsection{MPC Formulation}

For ease of exposition, we denote the dynamics model that combines the physics-based kinematics and GP-based dynamics developed in the previous section as
\begin{equation}
  \bbsym{x}_{k+1} = \FFF(\bbsym{x}_{k}, \bbsym{u}_{k}, \alpha_{k}, \gamma_{k}).
\end{equation}
We consider an MPC problem with a control horizon of length $H$, where the vehicle is required to track a given reference trajectory. The objective function is
\begin{equation}
\label{eq:obj}
\begin{multlined}
J (\bbsym{u}_{\cdot \mid k}, \bbsym{x}_{\cdot \mid k}) =\sum_{t=0}^{H-1} \Big( ||\bbsym{x}_{k+t|k} - \bbsym{x}^{\rm ref}_{k+t|k}||_{\bbsym{Q}}^2 + ||\bbsym{u}_{k+t|k}||_{\bbsym{R}}^2 \Big) \\
\!\! + \sum_{t=0}^{H-2}||\bbsym{u}_{k+t|k} - \bbsym{u}_{k+t+1|k}||_{\bbsym{R_d}}^2 
+ ||\bbsym{x}_{k+H|k}-\bbsym{x}^{\rm ref}_{k+H|k}||_{\bbsym{Q}_H}^2,
\end{multlined}
\end{equation}
where $\bbsym{x}_{\cdot \mid k} = [\bbsym{x}_{k|k}, \dots, \bbsym{x}_{k+H|k}]$ is the predicted state trajectory, 
$\bbsym{u}_{\cdot \mid k} = [\bbsym{u}_{k|k}, \dots, \bbsym{u}_{k+H-1|k}]$ is the control input sequence over the control horizon,
and $\bbsym{x}^{\rm ref}_{\cdot \mid k} = [\bbsym{x}_{k|k}^{\rm ref}, \dots, \bbsym{x}_{k+H|k}^{\rm ref}]$ is the reference trajectory, which consists of reference positions in global coordinates ($p_x$, $p_y$), velocity, and orientation.
The matrices $\bbsym{Q}$, $\bbsym{Q}_T$, $\bbsym{R}$, and $\bbsym{R_d}$ are weighting matrices.
The objective function \eqref{eq:obj} penalizes the deviation of the vehicle trajectory from the reference trajectory, the magnitude of the control inputs, and the variation of the control inputs over time.

Therefore, at every time step $k$, given the current state $\bbsym{x}_k$, we solve the following MPC problem:
\begin{subequations}
\label{eq:NP_MPC}
\begin{align}
\minimize_{\bbsym{u}_{\cdot|k}, \bbsym{x}_{\cdot|k}} \quad & 
J (\bbsym{u}_{\cdot \mid k}, \bbsym{x}_{\cdot \mid k}),
\label{equation/cost}\\
\textrm{subject to} \quad &\bbsym{x}_{k|k} = \bbsym{x}_k,\\[2pt]
& 
\begin{multlined}
\bbsym{x}_{k+t+1|k} = \FFF\big(\bbsym{x}_{k+t|k}, \bbsym{u}_{k+t|k}, \alpha_{k+t|k}, \gamma_{k+t|k}\big),
\end{multlined}
\\
& \alpha_{k+t|k} = \alpha\big(p_{x,k+t|k}, p_{y,k+t|k}\big), \\
& \gamma_{k+t|k} = \gamma\big(p_{x,k+t|k}, p_{y,k+t|k}\big), \\
& \bbsym{u}_{k+t|k} \in \mathcal{U}, \label{equation/constraint1} \\ 
& \forall t \in \{0, 1, \dots, H-1\},
\end{align}
\end{subequations}
where $\mathcal{U}$ is the set of admissible control inputs.
The MPC problem \eqref{eq:NP_MPC} is implemented in a receding horizon manner, in which the optimization is solved at every time step, and only the first control input is applied to control the vehicle to the next time step.

\subsection{Trajectory Rollouts and MPPI}
\label{subsec:rollouts}
Since the nonplanar vehicle model is highly nonlinear, we solve the nonlinear MPC problem \eqref{eq:NP_MPC} using Model Predictive Path Integral (MPPI) control, a sampling-based MPC method~\cite{williams2017information}. We now describe how the learning-based nonplanar vehicle model is integrated into the computation of the trajectory rollouts.

To compute the predicted state trajectory, we first initialize $\bbsym{x}_{k|k} = \bbsym{x}_{k}$. Then, at each step $t$ of the control horizon, given the current predicted vehicle position $(p_{x,k+t|k}, p_{y,k+t|k})$, we compute the roll and pitch angles $\alpha_{k+t|k}$ and $\gamma_{k+t|k}$ at that location from the normal-interpolating grids described in Section~\ref{sec:terrain_modeling}. Using these angles, we propagate the state forward to $\bbsym{x}_{k+t+1|k}$ with the learning-based vehicle dynamics from Section~\ref{sec:model-b}. This procedure is repeated until the end of the control horizon. At each horizon step, we interpolate $(\alpha, \gamma)$ from the terrain grid at the predicted position at propagate the state with $\FFF(\cdot)$. Repeating this over the horizon incorporates look-ahead terrain geometry into the dynamics.

At each iteration of MPPI, we first sample $N$ control input sequences $\bbsym{u}^{(s)}_{\cdot \mid k},\ s=1,\ldots, N$, from a \textit{truncated} normal distribution
\[
\mathcal{N}_{\text{trunc}}\bigl(\bbsym{u}^{\star}_{\cdot \mid k-1}, \bbsym{\Sigma}, \mathcal{U}\bigr),
\]
where $\bbsym{u}^{\star}_{\cdot \mid k-1}$ is the mean of the \textit{parent} normal distribution and the optimal control from the previous time-step, $\bbsym{\Sigma}$ is its user-prescribed covariance, and $\mathcal{U}$ specifies the truncation range so that the input constraints~\eqref{equation/constraint1} are satisfied. Then, starting from $\bbsym{x}_k$, we roll out the dynamics for each sampled control input sequence over the horizon $\bbsym{u}^{(s)}_{k+t \mid k}$ using the learning-based model $\FFF(\cdot)$ and the elevation map to obtain $\bbsym{x}^{(s)}_{k+t+1 \mid k}$. In practice, sampling of the control sequences $\bbsym{u}^{(s)}_{\cdot \mid k}$, rolling out the dynamics, and interpolating the roll and pitch angles are all executed in parallel on a GPU. 

We then perform \textit{importance sampling} on $\bbsym{u}^{(s)}_{\cdot \mid k}$ to obtain the approximate optimal predicted control sequence as
\begin{equation} 
\bbsym{u}^{\star}_{k+t \mid k} = \sum_{s=1}^N w_s \,\bbsym{u}^{(s)}_{k+t \mid k}, \; t \in \{0, \ldots, H-1\},
\label{eq:importance_sampling_1}
\end{equation}
with importance sampling weights $w_s$ given by
\begin{equation}
w_s = \frac{\exp\!\left(-\frac{1}{\tau}\, J\bigl(\bbsym{u}^{(s)}_{\cdot \mid k}, \bbsym{x}^{(s)}_{\cdot \mid k}\bigr) \right)}
{\sum_{j=1}^N \exp\!\left(-\frac{1}{\tau}\, J\bigl(\bbsym{u}^{(j)}_{\cdot \mid k}, \bbsym{x}^{(s)}_{\cdot \mid k}\bigr) \right)},
\label{eq:importance_sampling_2}
\end{equation}
where $\tau>0$ is a temperature parameter that tunes the sharpness of importance sampling. In our implementation, we use a horizon of $H = 20$ steps, $N = 1024$ sampled trajectories, temperature $\tau = 0.01$, and control input covariance $\bbsym{\Sigma} = \sigma_u^2 \mathbf{I}$ with $\sigma_u = 0.5$.

\section{Results and Discussions}
\label{sec:sim}

In this section, we validate the performance of the proposed control framework in a simulation environment on different generated nonplanar tracks. 

\subsection{Simulation Setup and Implementation Details}

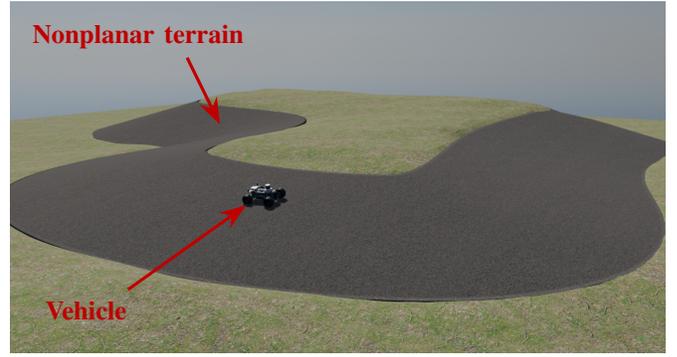
\begin{figure}[t]
\centering
\begin{tikzpicture}
  \node[anchor=south west, inner sep=0] (img) at (0,0)
    {\includegraphics[width=0.48\textwidth,trim=150 100 100 20,clip]{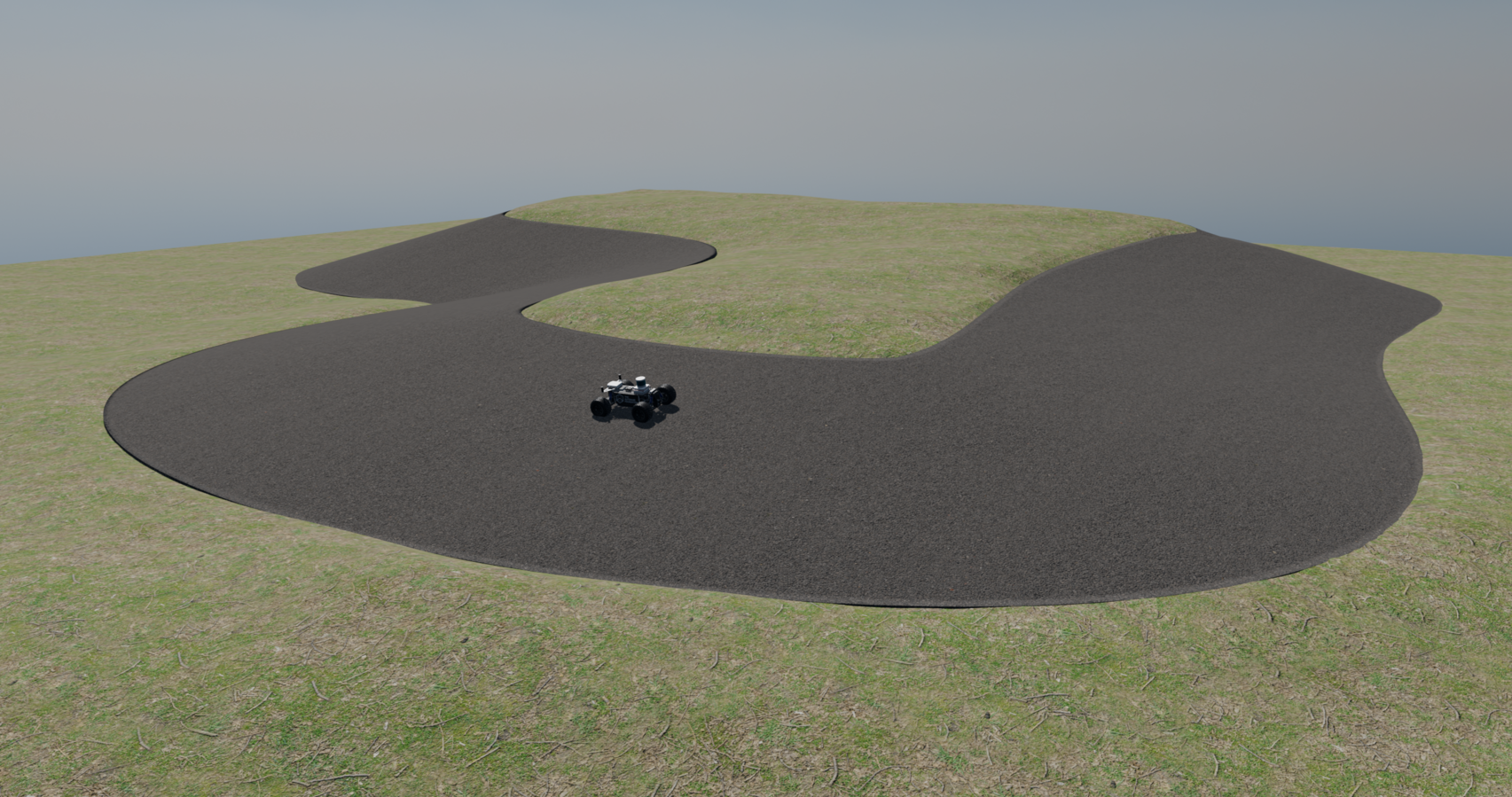}}; 
  \begin{scope}[x={(img.south east)}, y={(img.north west)}]
    \draw[draw={rgb,255:red,192; green,0; blue,0}, line width=1.2pt, -{Stealth[length=3.5mm,width=2.5mm]}]
      (0.27,0.84) -- (0.32,0.65);
    \node[anchor=west, text={rgb,255:red,192; green,0; blue,0}, font=\bfseries\normalsize]
      at (0.02,0.90) {Nonplanar terrain};

    \draw[draw={rgb,255:red,192; green,0; blue,0}, line width=1.2pt, -{Stealth[length=3.5mm,width=2.5mm]}]
      (0.18,0.18) -- (0.36,0.42);
    \node[anchor=west, text={rgb,255:red,192; green,0; blue,0}, font=\bfseries\normalsize]
      at (0.04,0.12) {Vehicle};
  \end{scope}
\end{tikzpicture}
\caption{The simulation environment for autonomous nonplanar vehicle navigation in \texttt{Isaac Sim}.}
\label{fig:isaacsim}
\vspace{-6mm}
\end{figure}

\begin{figure*}[t]
  \centering
  \begin{subfigure}[b]{0.32\textwidth}
    \includegraphics[width=1.0\textwidth]{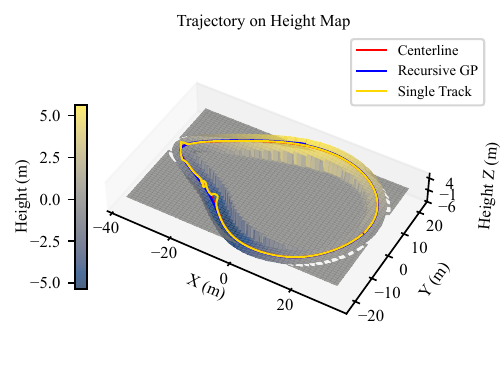}
  \end{subfigure}
  \hfill
  \begin{subfigure}[b]{0.32\textwidth}
    \includegraphics[width=1.0\textwidth]{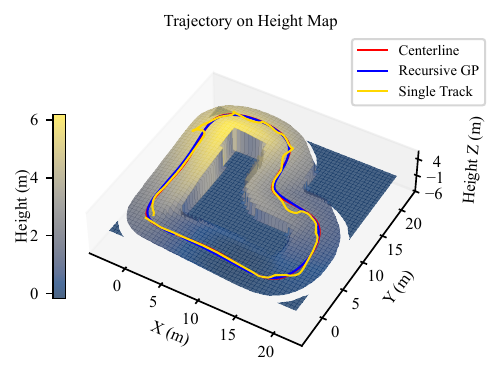}
  \end{subfigure}
  \hfill
  \begin{subfigure}[b]{0.32\textwidth}
    \includegraphics[width=1.0\textwidth]{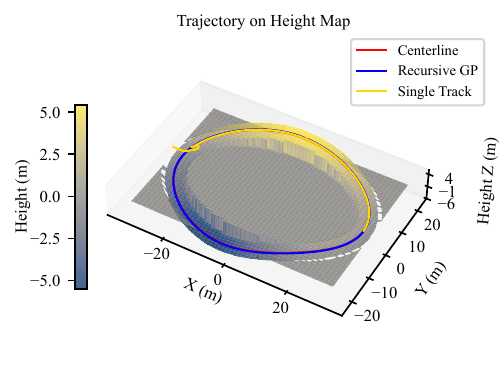}
  \end{subfigure}
  \vspace{-4mm}

  \begin{subfigure}[b]{0.32\textwidth}
    \includegraphics[width=0.96\textwidth]{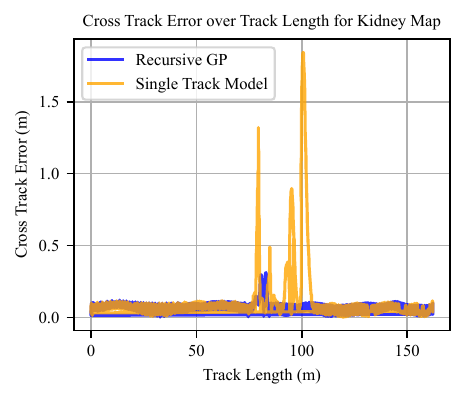}
    \vspace{-5pt}
    \caption{Kidney-shaped track}\label{fig:traject-a}
  \end{subfigure}
  \hfill
  \begin{subfigure}[b]{0.32\textwidth}
    \includegraphics[width=0.96\textwidth]{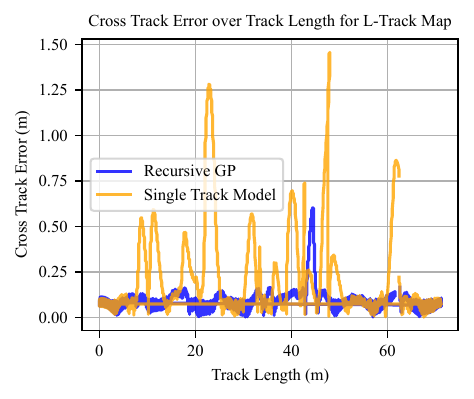}
    \vspace{-5pt}
    \caption{L-shaped track}\label{fig:traject-b}
  \end{subfigure}
  \hfill
  \begin{subfigure}[b]{0.32\textwidth}
    \includegraphics[width=0.96\textwidth]{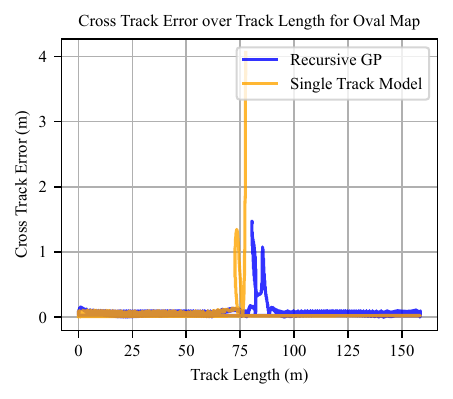}
    \vspace{-5pt}
    \caption{Oval track}\label{fig:traject-c}
  \end{subfigure}
  
  \caption{Top panels: Trajectories of the vehicle on different nonplanar tracks using a single-track model and the proposed Recursive GP. Bottom panels: Cross-track errors of the vehicle trajectories in simulation using the two different models.}
  \label{fig:traject}
  \label{fig:cte}
  \vspace{-6mm}
\end{figure*}

To simulate nonplanar vehicle dynamics and validate the proposed control framework, we used \texttt{NVIDIA Isaac Sim} \cite{nvidia_isaacsim}. 
We developed a custom simulation environment using \texttt{Isaac Sim}, which includes the 3D maps with varying geometry and a vehicle model with ROS2 support for control and publishing the ground-truth states. 
The environment is created using \texttt{Isaac Sim}'s Python API, where we load in a pre-made 3D track model's mesh after converting it to USD format. 
For the vehicle model, we use the built-in Leatherback car model, which is a 4-wheeled vehicle with Ackermann steering based on the \texttt{F1tenth} platform~\cite{okelly2020f1tenth}. 
The simulator publishes the vehicle's ground-truth states and accepts control commands over ROS2 topics. 
We implement the proposed nonplanar MPC framework as a separate ROS2 node, which subscribes to the vehicle's ground-truth states and publishes the control commands. To facilitate reproducibility and reuse of the simulation environment, we open-source our code at \url{https://github.com/mlab-upenn/NonPlanarIsaacSim}.
The simulation environment in \texttt{Isaac Sim} is illustrated in Fig.~\ref{fig:isaacsim}. To control the vehicle using an MPPI controller with a single-track dynamic bicycle model, we identify the ground-truth parameters of the vehicle from Isaac Sim. These include the wheelbase, the center of gravity, inertia, mass, and friction. The cornering stiffness of the linear tire model was set to be the same as that of the F1tenth platform as an empirical estimate.

\subsection{Results}

\begin{figure*}[t]
\centering
\begin{subfigure}{0.33\textwidth}
\centering
\includegraphics[width=0.96\textwidth,trim=0 0 0 0,clip]{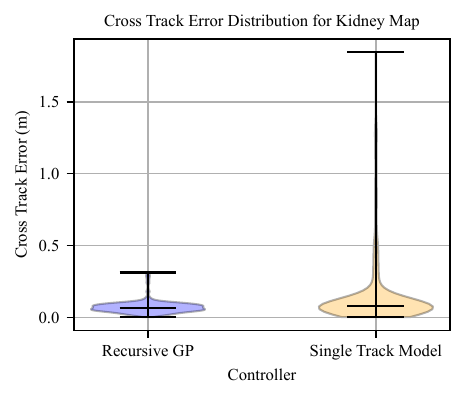}
\end{subfigure}
\hspace{-30mm}
\hfill
\begin{subfigure}{0.33\textwidth}
\centering
\includegraphics[width=0.96\textwidth,trim=0 0 0 0,clip]{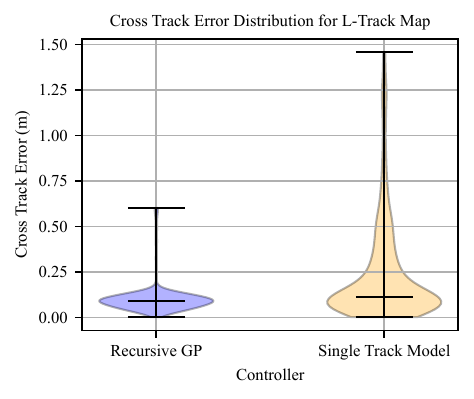}
\end{subfigure}
\hspace{-30mm}
\hfill
\begin{subfigure}{0.33\textwidth}
\centering
\includegraphics[width=0.96\textwidth,trim=0 0 0 0,clip]{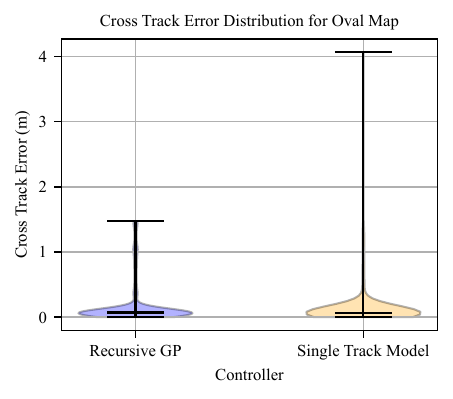}
\end{subfigure}
\vspace{-5pt}
\caption{Distributions of absolute cross-track errors using the recursive GP residual model and the single-track dynamic model.}
\label{fig:cte_histogram}
\vspace{-4mm}
\end{figure*}

\begin{figure*}[t]
\centering
\begin{subfigure}{0.33\textwidth}
\centering
\includegraphics[width=0.96\textwidth,trim=0 0 0 0,clip]{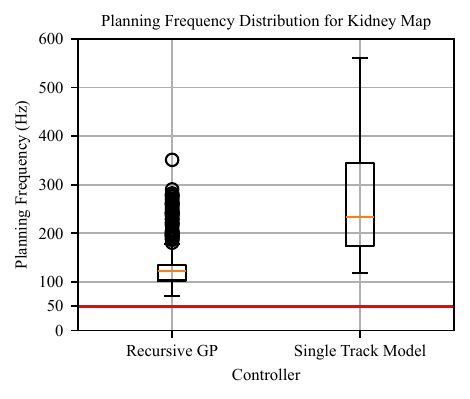}
\end{subfigure}
\hspace{-30mm}
\hfill
\begin{subfigure}{0.33\textwidth}
\centering
\includegraphics[width=0.96\textwidth,trim=0 0 0 0,clip]{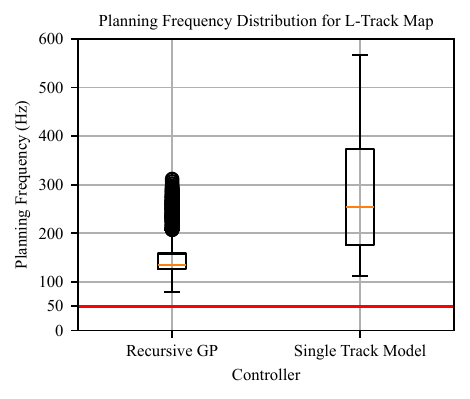}
\end{subfigure}
\hspace{-30mm}
\hfill
\begin{subfigure}{0.33\textwidth}
\centering
\includegraphics[width=0.96\textwidth,trim=0 0 0 0,clip]{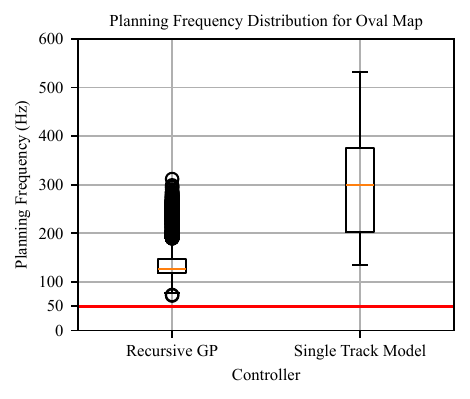}
\end{subfigure}
\caption{Distributions of control frequencies using the recursive GP residual model and the single-track dynamic model. The minimum desired frequency of 50Hz is shown in red.}
\label{fig:freq_histogram}
\vspace{-6mm}
\end{figure*}

We created three different nonplanar tracks, namely kidney, L-shaped, and oval, in the \texttt{Isaac Sim} environment for validation. The statistics for each of the three tracks can be seen in Table~\ref{tab:map_statistics}.

\begin{table}[htbp]
    \centering
    \caption{Map Statistics}
    \begin{tabular}{lrrr}
        \toprule
        Metric & Kidney & L-Track & Oval \\
        \midrule
        Min elevation (m) & -5.389 & -0.183 & -5.523 \\
        Max elevation (m) & 5.696 & 6.234 & 5.538 \\
        Elevation range (m) & 11.085 & 6.416 & 11.061 \\
        Max slope (degrees \degree) & 27.811 & 75.760 & 23.177 \\
        Median slope (degrees \degree) & 10.051 & 14.698 & 11.819 \\
        \bottomrule
    \end{tabular}
    \label{tab:map_statistics}
\end{table}
\vspace{-3mm}

We compared the tracking control performance of the MPPI using the learning-based model with the recursive sparse GP, with a baseline MPPI using the baseline single-track dynamic model without residual compensation.  
We show the vehicle trajectories on the nonplanar tracks in Fig~\ref{fig:traject}.
The figure shows that, for kidney and L-shaped tracks (Figures~\ref{fig:traject-a} and~\ref{fig:traject-b}), the MPPI controller using either model can complete the lap, but the single-track dynamic model results in a lot of fluctuation around the reference trajectory, while the learning-based model enables MPPI to track the reference trajectory closely.
Meanwhile, for the oval track (Fig.~\ref{fig:traject-c}), using MPPI with the single-track dynamic model, the baseline model fails to maintain stability, resulting in track departure, whereas the proposed framework preserves traversability despite the high-curvature slopes.

To better evaluate tracking performance, we show in Fig.~\ref{fig:cte} the cross-track errors for the comparison between the controller using the single-track baseline model and using the learning-based model with recursive sparse GP.
Additionally, we show the distributions of the cross-track absolute errors in the form of histograms in Fig.~\ref{fig:cte_histogram}.
Overall, using the residual dynamic model in addition to the single-track model can result in lower cross-track errors compared to using purely the single-track model.
Therefore, the residual model can compensate for the mismatch between the single-track planar model and the actual nonplanar vehicle dynamics.

Finally, to compare computational complexity between the proposed learning-based model and the baseline single-track model, we report control frequencies, computed as the inverse of the MPPI solve time. Integrating the sparse GP residual model increases computation time, as trajectory rollouts require GP predictions at each time step over the control horizon. However, note that even with the use of a complex GP-based model, the computation time generally is no longer than $\SI{20}{ms}$ (or \SI{50}{Hz}), which is highly practical for real-time control. 

\section{Conclusions}
\label{sec:cls}

This paper presents a control framework for autonomous driving on nonplanar terrain that combines Model Predictive Path Integral (MPPI) control with a recursively updated sparse Gaussian process (GP) dynamics model. The vehicle dynamics are represented as a nominal single-track model plus a sparse GP residual that is learned online to capture time-varying surface geometry. The learned dynamics model is embedded in a model predictive control formulation, and MPPI is used to solve the resulting nonlinear optimal control problem. To evaluate the framework, we developed and open-source a simulation environment in NVIDIA Isaac Sim using several nonplanar tracks. The results show that the recursively updated sparse GP residual model significantly improves tracking performance compared to using the nominal single-track model alone. In the future, we plan to conduct physical experiments on real-world off-road vehicles and incorporate dynamic obstacles into the nonplanar MPPI framework.

\bibliographystyle{IEEEtran}
\bibliography{IEEEabrv,references}

\end{document}